\documentclass[sigconf,natbib=true]{acmart}

\setcopyright{none}
\renewcommand\footnotetextcopyrightpermission[1]{}
\acmConference[~\img{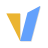} vLLM Semantic Router Project]{}{\href{https://github.com/vllm-project/semantic-router}{\img{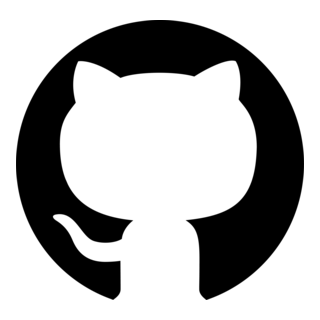}}}{\href{https://huggingface.co/llm-semantic-router}{\img{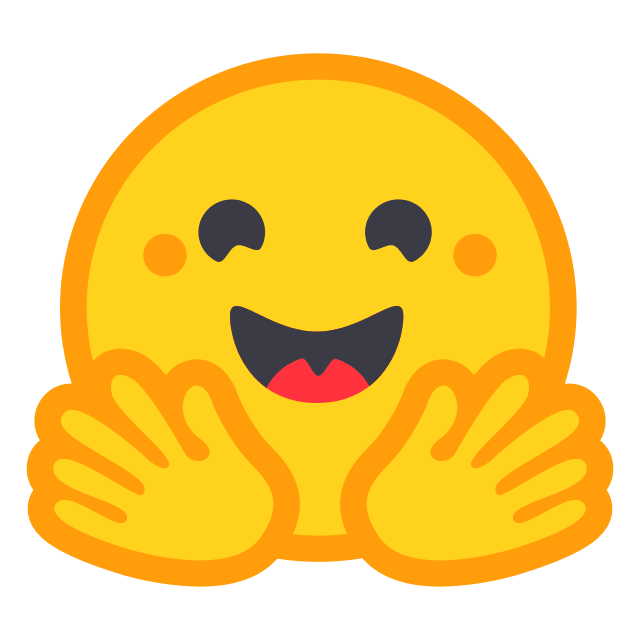}}}

\usepackage{booktabs}
\usepackage{tikz}
\usetikzlibrary{shapes,arrows,positioning}
\usepackage{fancyhdr}
\usepackage{graphicx}
\newcommand*{\img}[1]{%
    \raisebox{-.01\baselineskip}{%
        \includegraphics[
        height=0.9\baselineskip,
        width=0.9\baselineskip,
        keepaspectratio,
        ]{#1}%
    }%
}

\title{Fast and Faithful: Real-Time Verification for Long-Document Retrieval-Augmented Generation Systems}

\author{Xunzhuo Liu}
\affiliation{%
  \institution{vLLM Semantic Router Project}
  \country{United States}}
\email{mixdeers@gmail.com}

\author{Bowei He}
\authornote{Corresponding author}
\affiliation{%
  \institution{MBZUAI, McGill University}
  \country{UAE, Canada}}
\email{Bowei.He@mbzuai.ac.ae}

\author{Xue Liu}
\affiliation{%
  \institution{MBZUAI, McGill University}
  \country{UAE, Canada}}
\email{steve.liu@mbzuai.ac.ae}

\author{Haichen Zhang}
\affiliation{%
  \institution{AMD}
  \country{United States}}
\email{haichen.zhang@amd.com}

\author{Huamin Chen}
\affiliation{%
  \institution{Red Hat}
  \country{United States}}
\email{hchen@redhat.com}

\begin{abstract}
Retrieval-augmented generation (RAG) is increasingly deployed in enterprise search and document-centric assistants, where responses must be grounded in long and complex source materials. In practice, verifying that generated answers faithfully reflect retrieved documents is difficult: large language models can check long contexts but are too slow and costly for interactive services, while lightweight classifiers operate within strict context limits and frequently miss evidence outside truncated passages. We present the design of a real-time verification component integrated into a production RAG pipeline that enables full-document grounding under latency constraints. The system processes documents up to 32K tokens and employs adaptive inference strategies to balance response time and verification coverage across workloads. We describe the architectural decisions, operational trade-offs, and evaluation methodology used to deploy the verifier, and show that full-context verification substantially improves detection of unsupported responses compared with truncated validation. Our experience highlights when long-context verification is necessary, why chunk-based checking often fails in real documents, and how latency budgets shape model design. These findings provide practical guidance for practitioners building reliable large-scale retrieval-augmented applications. \footnote{Model, benchmark, and code: \url{https://huggingface.co/llm-semantic-router}}
\end{abstract}



\begin{document}

\maketitle

\section{Introduction}
Retrieval-augmented generation (RAG) has become a standard approach for grounding large language model (LLM) outputs in external documents~\cite{lewis2020retrieval}, like legal contracts~\cite{wiratunga2024cbr}, financial reports~\cite{setty2024improving}, and scientific papers~\cite{yu2025rag}. It is now widely deployed in enterprise search and document-centric assistants. In production systems, however, retrieving relevant passages is only part of the problem. A system must also verify that a generated answer is actually supported by the retrieved sources under strict latency and cost constraints~\cite{filice2025generate}. This verification requirement is particularly important in compliance-sensitive scenarios such as contract analysis, clinical reporting, and financial documentation, where unsupported claims may carry real operational risk.

A practical obstacle is document length. Real-world documents are often far longer than the context limits of lightweight verification models: contracts frequently span 15K--30K tokens, clinical trial reports 20K--50K tokens, and regulatory filings may exceed 100K tokens. Most deployed verification components rely on encoder-based classifiers with context windows around 8K tokens~\cite{zimmerman2024two}. As a result, verification typically becomes truncated validation, where the system checks only a prefix or a small subset of passages rather than the full document. In realistic workflows, critical evidence may appear far from the beginning of a document. For example, if an answer claims that a contract permits termination with 30 days notice, the relevant clause may appear deep within the file. A short-context verifier never observes this clause and therefore cannot reliably determine whether the claim is supported.

This limitation produces a persistent \textbf{speed–context tradeoff}. Long-context LLM-as-judge approaches~\cite{manakul2023selfcheckgpt,min2023factscore} can examine full documents, but their latency and cost are unsuitable for interactive services. Encoder-based verifiers~\cite{wu2024ragtruth,kovacs2025lettuce} are fast and inexpensive, yet their limited context leads to incomplete verification coverage. Production systems often attempt to bridge the gap using chunking and aggregation heuristics, but these approaches fail when evidence spans multiple sections or depends on global document structure, and they introduce ambiguity when chunk-level decisions disagree. In practice, systems must choose between slow but reliable verification and fast but incomplete verification.

We address this problem by extending an encoder-based verifier to operate on full documents while maintaining real-time inference. Achieving this capability required more than simply enlarging a context window. Existing hallucination detectors are trained and evaluated on short passages, and their behavior changes significantly when exposed to full-length documents containing irrelevant sections, repeated entities, and long-distance dependencies. As a result, both the faithfulness verification model and its evaluation setting must be redesigned for long-document verification. Technically, the main obstacle lies in extending rotary position embeddings (RoPE)~\cite{su2021roformer} to bidirectional encoders. Although RoPE scaling has enabled long contexts in decoder-only LLMs~\cite{peng2023yarn}, directly applying it to encoders degrades long-range sensitivity. Because encoders attend to all positions simultaneously, naive long-context masked-language-model training overwrites attention patterns that retrieval-sensitive tasks depend on. We therefore develop a retrieval-aware RoPE extension strategy and train a long-context hallucination detector capable of processing entire documents. To support deployment, we further introduce adaptive early-exit inference so the verifier can adjust computation based on document length and latency budget. Together, these components enable full-document grounding for long inputs while preserving throughput on short queries, and require a corresponding long-document evaluation setting to measure verification reliability in realistic workloads.

Our contributions are summarized as follows:
\begin{enumerate}
\item We develop a retrieval-aware RoPE extension method that scales encoder context windows to 32K tokens while preserving long-range retrieval sensitivity.
\item We train a long-context hallucination detector and identify practical fine-tuning strategies important for deployment.
\item We design a configurable early-exit inference framework enabling explicit latency–accuracy tradeoffs suitable for heterogeneous workloads.
\item We construct a long-document evaluation setting demonstrating the necessity of full-context verification compared to truncated baselines.
\end{enumerate}

\section{Methodology}
\label{sec:method}

\subsection{Overview}
\label{sec:method_overview}
We study \textit{long-context hallucination detection} for retrieval-augmented generation.
Given a retrieved evidence context, a user query, and a model response, our verifier predicts which response tokens are unsupported by the provided evidence.
Our methodology has three components:
(1) extending an encoder to long context via RoPE scaling while preserving long-range attention;
(2) training a long-context hallucination detector with response-only supervision and deployment-oriented fine-tuning choices;
(3) enabling configurable inference cost via early-exit adapters for production-scale serving.

\subsection{Long-Context Encoder Extension}
\label{sec:extension}

%\textbf{Background: RoPE scaling for encoders.}
ModernBERT-style encoders~\cite{warner2025smarter} use Rotary Position Embeddings (RoPE), where attention between positions depends on relative distance via rotation matrices.
Because the encoder is pre-trained with a fixed maximum position, extrapolating beyond the pre-training length can yield unstable behavior.
We adopt YaRN~\cite{peng2023yarn} to extrapolate RoPE by interpolating rotation frequencies:
\begin{equation}
\theta'_i = \theta_i \cdot s^{-1} \quad \text{for low frequencies},
\end{equation}
where $s$ is a scaling factor that increases the supported context length. A straightforward approach is to apply RoPE scaling and then fine-tune on long documents with standard masked language modeling (MLM).
However, this can cause the encoder to forget long-range attention patterns learned during pre-training, harming long-distance evidence utilization. We view this as an optimization pathology rather than an architectural limitation.

%\noindent \textbf{Why standard MLM under-trains long-range attention.}
Standard MLM often provides weak gradients for long-distance dependencies~\cite{joshi2020spanbert, zhang2020pegasus, wu2025longattn} due to:
(1) \textit{local sufficiency}, where masked tokens can be predicted from nearby context;
(2) \textit{optimization shortcuts}, where ignoring distant tokens minimally increases loss;
(3) \textit{weight overwriting}, where fine-tuning updates disrupt pre-trained attention circuits.

\begin{figure}[t]
  \centering
  \includegraphics[width=0.48\textwidth]{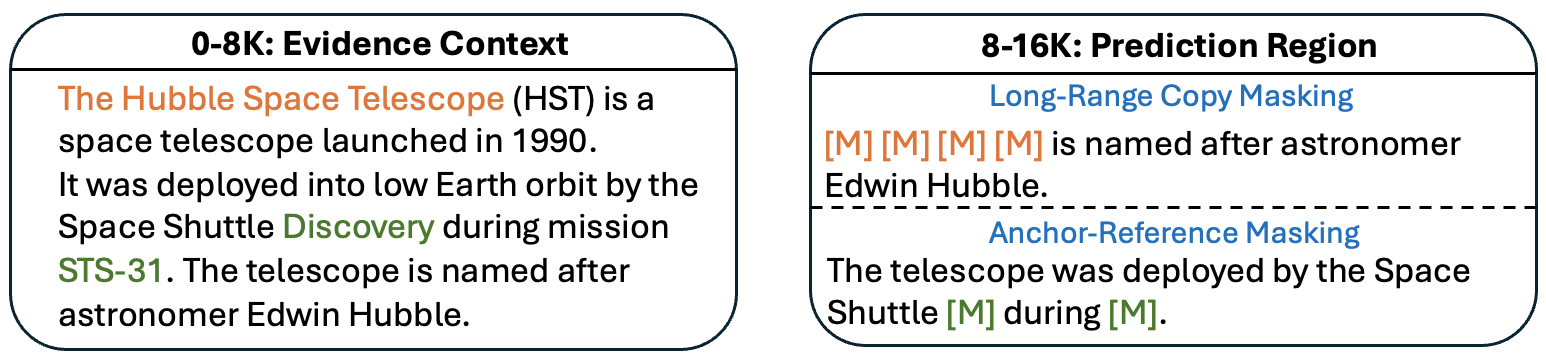}
  %\vspace{-3mm}
  \caption{Retrieval-aware masking. Tokens are masked in the second half such that prediction requires attending across long distances to earlier evidence.}
  \label{fig:retrieval_masking}
\end{figure}

%\noindent \textbf{Retrieval-aware masking.}
To explicitly train and preserve long-range attention, we introduce \textbf{retrieval-aware masking} strategies (see Figure~\ref{fig:retrieval_masking}) that require attending across large distances to recover masked tokens: 1) \textit{Long-Range Copy Masking}: within each long sequence, identify tokens that appear in both the first and second halves, and mask occurrences in the second half so correct prediction requires attending to the distant earlier mention; 2) \textit{Anchor-Reference Masking}: select a small set of early anchor positions and mask later occurrences of anchor tokens, forcing attention back to anchors. These masks inject direct gradient signals into long-range pathways that standard MLM tends to ignore.

%\vspace{-1mm}

%\noindent \textbf{Elastic weight consolidation.}
To reduce destructive drift from pre-trained parameters, we regularize updates with Elastic Weight Consolidation (EWC)~\cite{kirkpatrick2017overcoming}:
\begin{equation}
\mathcal{L}_{total} = \mathcal{L}_{MLM} + \frac{\lambda}{2} \sum_i F_i (\theta_i - \theta_i^*)^2,
\end{equation}
where $\theta^*$ are pre-trained weights and $F_i$ is the Fisher information measuring parameter importance. This penalizes changes to weights that are critical for the pre-trained representation. In addition to regularization, we adopt a conservative fine-tuning regime to avoid overwriting pre-trained long-range circuits (e.g., small-step updates and limited exposure, i.e., low learning rates and single epoch setting), which we find complements retrieval-aware masking and EWC.

\subsection{Hallucination Detection Fine-tuning}
\label{sec:detector}
\begin{figure}[t]
  \centering
  \includegraphics[width=0.46\textwidth]{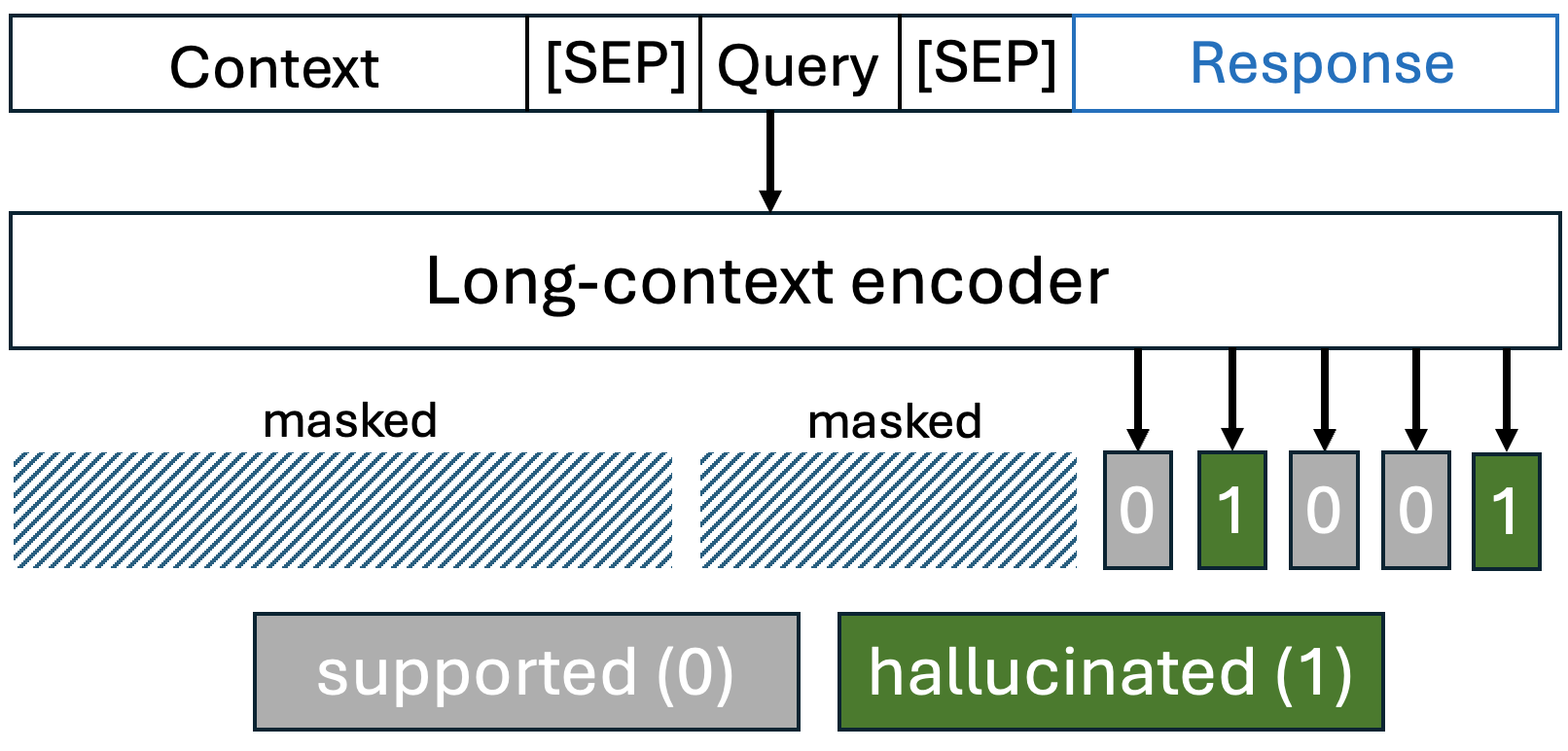}
  \vspace{-3mm}
  \caption{Hallucination detection fine-tuning with response-only supervision.}
  \label{fig:response_only_supervision}
\end{figure}

We fine-tune the long-context encoder as a hallucination detector via token-level classification on a concatenated input:
\begin{equation}
\text{Input} = [\text{Context}] \oplus [\text{SEP}] \oplus [\text{Query}] \oplus [\text{SEP}] \oplus [\text{Response}].
\end{equation}
As shown in Figure~\ref{fig:response_only_supervision}, the model predicts a binary label for each \textit{response} token: supported (0) or hallucinated (1).
Tokens belonging to the context and query are excluded from the loss (masked labels), ensuring supervision targets faithfulness judgments on the generated content. We optimize a standard token-level classification loss over response tokens. At inference time, token predictions can be aggregated into span-level hallucination highlights (contiguous hallucinated tokens) and an example-level decision (e.g., any hallucinated span present), enabling both UX-facing explanations and system-level gating.

%\vspace{-3mm}

%\paragraph{Practical fine-tuning strategies for deployment.}
We found several fine-tuning choices materially affect deployability and reliability: 1) \textit{Prefer simple losses over imbalance-specific objectives.} Although hallucinations are relatively sparse at the token level, we found that imbalance-focused losses (e.g., focal loss or heavily class-weighted cross-entropy) tend to over-predict hallucinations, boosting recall at the expense of precision. In deployment, this leads to excessive false alarms and poor user experience. A standard cross-entropy objective provides a more stable precision--recall trade-off; 2) \textit{Match the training distribution to the target RAG setting.} Fine-tuning benefits most from data whose hallucination rate and generation style resemble production RAG outputs. Data sources with near-constant hallucination (e.g., adversarial or synthetic QA-style sets) can bias the detector toward pessimistic predictions and reduce overall F1. In contrast, distribution-matched data improves robustness under realistic hallucination prevalence; 3) \textit{Keep supervision localized to the response.} Masking context/query tokens in the loss focuses capacity on faithfulness judgments rather than learning to ``label'' evidence itself. This also avoids degenerate solutions where the model learns dataset artifacts in the context portion; 4) \textit{Stabilize training for long contexts.} Long-context inputs increase memory and reduce effective batch size, which can make optimization noisier. In practice we prioritize stable training dynamics (e.g., avoiding aggressive objectives that amplify rare-token gradients) to reduce run-to-run variance and improve reproducibility in production pipelines.

%\paragraph{Compatibility with long-context evidence placement.}
Because evidence for (or against) a claim may appear anywhere in a long document, the detector is trained to condition on the entire retrieved context rather than chunk-local windows. This aligns the model with production RAG failure modes, where truncation or partial context can silently eliminate crucial evidence.

\subsection{Configurable Early-Exit Inference}
\label{sec:production_method}

\begin{figure}[t]
  \centering
  \includegraphics[width=0.48\textwidth]{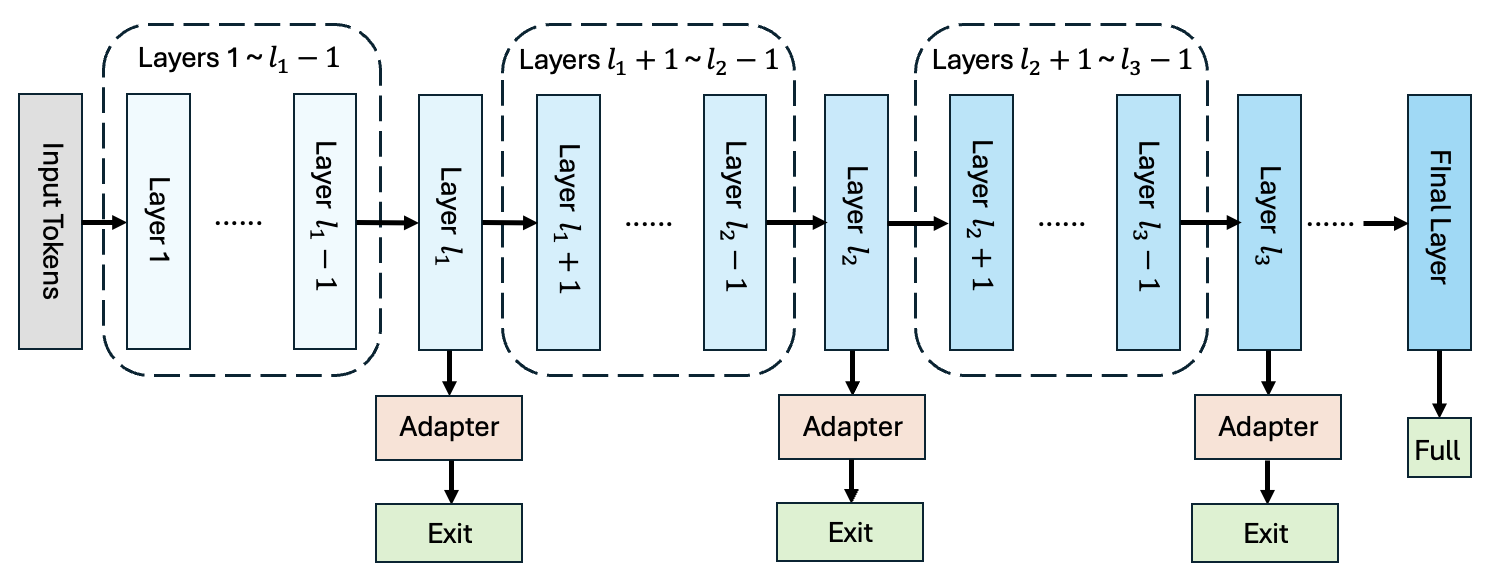}
  \caption{Early-exit architecture. Lightweight adapters at intermediate layers enable configurable compute.}
  \label{fig:early_exit}
\end{figure}

%\noindent \textbf{Motivation.}
Running the full transformer stack over long contexts is expensive~\cite{xin2020deebert, liu2020fastbert}.
We therefore enable configurable compute by allowing the verifier to exit early at intermediate layers, trading accuracy for latency when needed. As shown in Figure~\ref{fig:early_exit}, we attach lightweight classifier adapters at selected intermediate layers $l$ (e.g., $l_1$, $l_2$, $l_3$ in the figure).
Each adapter maps hidden states $\mathbf{h}_l$ to token labels:
\begin{equation}
\text{Adapter}_l = \text{Linear}(\text{GELU}(\text{Linear}(\text{LayerNorm}(\mathbf{h}_l)))).
\end{equation}
This design adds minimal parameters and avoids modifying the backbone weights. We train intermediate adapters to match the predictions of the full-depth classifier on the same fine-tuned backbone (teacher), using a distillation objective. This yields strong early-exit performance without requiring separate end-to-end training for each exit point.

\vspace{-3mm}

\section{Experiments}
\label{sec:exp}
We evaluate four research questions: \textbf{RQ1:} Can the extended encoder preserve short-document performance while supporting long contexts? \textbf{RQ2:} Does full-context verification improve hallucination detection on long documents? \textbf{RQ3:} How do key design choices affect the hallucination detection performance? \textbf{RQ4:} Can early exit significantly reduce latency while maintaining detection quality?
\vspace{-2mm}

\subsection{Experiment Setup}
\subsubsection{Benchmarks}
\label{sec:datasets}
\textbf{Standard hallucination benchmark:} We evaluate short-context performance on the RAGTruth benchmark~\cite{wu2024ragtruth}, which contains human-annotated hallucination spans in RAG generated responses. The longest document in it is 2,632 tokens, well within traditional 8K limits. This benchmark primarily measures whether extending context length harms standard verification quality.
\textbf{Long-context benchmark:} Existing benchmarks do not require long-context reasoning. To evaluate full-document verification, we construct a long-context hallucination benchmark based on naturally long documents (See Table~\ref{tab:sources}) via following steps: (1) filter documents between 8K and 24K tokens, (2) generate responses using a large language model,
(3) inject hallucinated spans into 50\% of responses using controlled prompting, (4) annotate hallucination spans automatically and validate balance. The resulting test set contains 337 samples with an average length of $\sim$17.5K tokens.

\begin{table}[t]
\centering
\small
\caption{Source datasets of the long-context benchmark.}
\label{tab:sources}
\begin{tabular}{lccc}
\toprule
\textbf{Source} & \textbf{Domain} & \textbf{Token Range} & \textbf{Documents} \\
\midrule
NarrativeQA~\cite{kovcisky2018narrativeqa} & Stories & 8K--24K & 489 \\
QuALITY~\cite{pang2022quality} & Articles & 8K--20K & 223 \\
GovReport~\cite{huang2021efficient} & Government & 8K--20K & 88 \\
\bottomrule
\end{tabular}
\end{table}

\subsubsection{Evaluation Metrics} We report: 1) \textbf{Token F1:} token-level hallucination detection accuracy; 2) \textbf{Example F1:} whether a response contains hallucination; 3) \textbf{Hallucination Recall:} critical safety metric (missing hallucinations is costly).

\subsection{Experiment Results and Analysis}

\subsubsection{Short-Context Performance (\textbf{RQ1})}
\label{sec:short}
We first verify that long-context extension does not degrade standard verification tasks. From Table~\ref{tab:ragtruth_results}, despite supporting 4$\times$ longer context, our model achieves comparable performance to 8K baselines. This is important for deployment: a single model can replace short-context verifiers without performance loss.

\begin{table}[t]
\centering
\small
\caption{RAGTruth results. The 32K model preserves performance on short documents.}
\label{tab:ragtruth_results}
\begin{tabular}{lccc}
\toprule
\textbf{Model} & \textbf{Context} & \textbf{Token F1} & \textbf{Example F1} \\
\midrule
LettuceDetect-large & 8K & 0.6158 & 79.22\% \\
\textbf{Ours} & \textbf{32K} & 0.5337 & \textbf{77.00\%} \\
\bottomrule
\end{tabular}
\end{table}

\subsubsection{Long-Context Hallucination Detection (\textbf{RQ2})}
\label{sec:long}
We next evaluate full-document verification and results are provided in Table~\ref{tab:longcontext_results}. The 8K model misses most hallucinations because the evidence lies outside the visible window. When a 15K-token document is truncated to 8K, roughly half of the supporting evidence disappears. Our full-context verifier maintains access to the entire document and dramatically improves recall. We also observe that the performance gap increases with document length: the longer the document, the more severe truncation becomes for 8K models.
\begin{table}[t]
\centering
\small
\caption{Long-context hallucination detection. 8K models must truncate most evidence.}
\label{tab:longcontext_results}
\begin{tabular}{lccc}
\toprule
\textbf{Metric} & \textbf{32K Model} & \textbf{8K Model} & \textbf{Improvement} \\
\midrule
Samples Truncated & 0\% & 95\% & -- \\
Hallucination Recall & 0.55 & 0.06 & +817\% \\
\textbf{Hallucination F1} & \textbf{0.50} & 0.10 & \textbf{+400\%} \\
\bottomrule
\end{tabular}
\end{table}

\subsubsection{Ablation: Fine-tuning Strategies (\textbf{RQ3})}
\label{sec:finetune_ablation}
We study the fine-tuning choices in Section~\ref{sec:detector} on the RAGTruth validation set.
(1) \textit{Loss function.} Despite class imbalance, imbalance-aware objectives are detrimental. Focal loss ($\gamma\!=\!1,2$) and heavily weighted cross-entropy increase recall but sharply reduce precision through systematic over-prediction, labeling many supported tokens as hallucinated. Standard cross-entropy achieves the best F1 by maintaining a stable precision--recall trade-off, which is preferable in deployment where false positives are costly. (2) \textit{Training data distribution.} Increasing dataset size alone does not help. QA-style hallucination datasets have extremely high hallucination prevalence (89--100\%) compared to 43\% in RAGTruth, biasing the detector toward pessimistic predictions and lowering F1. In contrast, distribution-matched data-to-text datasets improve stability and detection quality, indicating that distribution alignment matters more than scale.

\subsubsection{Ablation: Long-Context Extension (\textbf{RQ3})}
\label{sec:rope_ablation}
We compare naïve long-document MLM fine-tuning with our retrieval-aware extension using a controlled retrieval task where a token must be predicted from distant evidence in the same sequence. After RoPE scaling and standard MLM fine-tuning, the encoder develops a strong locality bias: prediction accuracy remains high at short ranges but drops sharply once the required attention distance exceeds 1K tokens, indicating catastrophic forgetting of long-range behavior. With retrieval-aware masking and EWC, the model preserves long-distance attention, maintaining reliable retrieval up to roughly 2K tokens, where the naïve model already fails. Both approaches struggle beyond very long separations (e.g., $\geq$4K tokens). These results show that long-context capability is limited primarily by fine-tuning dynamics rather than positional extrapolation itself.

\subsubsection{Early Exit Evaluation (\textbf{RQ4})}
\label{sec:early_exit}
We probe hidden states across transformer layers using linear classifiers to identify where hallucination relevant features emerge. From Table~\ref{tab:early_exit}, we have several observations: Early layers already encode useful recall signals; Precision improves with depth; Intermediate layers provide strong detection features. Based on probing, we select three exit layers representing fast, balanced, and high-recall configurations. Especially, intermediate exits retain most accuracy while significantly reducing compute. Besides, from Table~\ref{tab:latency}, a key finding is that early-exit speedup \textit{increases} with context length. Skipping transformer layers avoids large attention computations, making early exit particularly valuable for long-document verification.

\begin{table}[t]
\centering
\small
\caption{Early exit performance on validation set.}
\label{tab:early_exit}
\begin{tabular}{lcccc}
\toprule
\textbf{Exit Layer} & \textbf{Example F1} & \textbf{\% of Full} & \textbf{Compute} & \textbf{Speedup} \\
\midrule
Full model (L22) & 95.5\% & 100\% & 100\% & 1.0$\times$ \\
Intermediate (L16) & 92.8\% & 97\% & 73\% & 1.4$\times$ \\
Mid (L11) & 81.2\% & 85\% & 50\% & 2.0$\times$ \\
Early (L6) & 48.2\% & 50\% & 27\% & 3.3--3.9$\times$ \\
\bottomrule
\end{tabular}
\end{table}

\begin{table}[t]
\centering
\small
\caption{Comparison of throughput (samples/sec) and latency.}
\label{tab:latency}
\begin{tabular}{lrrrr}
\toprule
\textbf{Sequence Length} & \textbf{BS=1} & \textbf{BS=4} & \textbf{BS=8} & \textbf{ms/sample} \\
\midrule
\multicolumn{5}{c}{\textit{Full Model (L22)}} \\
512 tokens & 62/s & 254/s & \textbf{487/s} & 2.1ms \\
8K tokens & 59/s & 86/s & 92/s & 10.9ms \\
16K tokens & 28/s & 32/s & 31/s & 32ms \\
24K tokens & 15/s & 16/s & 16/s & \textbf{63ms} \\
\midrule
\multicolumn{5}{c}{\textit{Early Exit (L16)}} \\
512 tokens & 86/s & 346/s & \textbf{685/s} & 1.5ms \\
16K tokens & 37/s & 43/s & 42/s & 24ms \\
24K tokens & 21/s & 21/s & 21/s & \textbf{47ms} \\
\bottomrule
\end{tabular}
\end{table}

\section{Conclusion and Future Works}
We present a real-time, full-context verification method for production long-document RAG systems. The approach extends the encoder context to 32K using retrieval-aware RoPE while preserving long-range capabilities, performs token-level hallucination detection to evaluate responses at the word level, and introduces configurable early-exit inference to achieve a controllable accuracy–throughput trade-off under varying document lengths and latency budgets. Experiments show that the method maintains near-baseline performance on short documents while significantly outperforming truncated validation on 8K–24K long-document verification. In the future, we plan to extend the system to longer contexts (64K–128K), enable confidence-aware dynamic inference for lower latency, and develop realistic multilingual long-document benchmarks to improve robustness and cross-domain reliability.

%\section*{Acknowledgments}

%We thank the Answer.AI team for ModernBERT and the authors of RAGTruth for their dataset.

\section*{Presenter Biography}
\textbf{Xunzhuo Liu}
is a systems and infrastructure engineer working at the intersection of large language models and networking. He co-founded and leads the vLLM Semantic Router, a system-level intelligence layer for Mixture-of-Models. He is a CNCF Ambassador, serves on the Envoy Gateway Steering Committee, and co-chairs the Kubernetes AI Gateway Working Group.

\noindent \textbf{Bowei He} is currently a postdoctoral researcher in MBZUAI and McGill University. He received the Ph.D. degree from City University of Hong Kong. His research focuses on information retrieval, language model, and agentic AI. He has published over 40 papers on top-tier venues including NeurIPS, ICLR, KDD, and WWW. 
\bibliographystyle{ACM-Reference-Format}
\bibliography{references}

\appendix

\section{Implementation Details}
\label{app:implementation}
We provide additional implementation details for long-context extension, hallucination detection fine-tuning, and early-exit adapter training.

For \textbf{long-context extension}, we start from the ModernBERT-base encoder~\cite{warner2025smarter} (149M parameters) and extend the positional range using YaRN-style~\cite{peng2023yarn} RoPE scaling with a $4\times$ expansion, increasing the maximum sequence length from 8K to 32K tokens. The model is trained with masked language modeling (MLM) using a masking probability of 0.30. To maintain sensitivity to long-range evidence within the sequence, we incorporate retrieval-aware masking with probability 0.10, which increases the likelihood that masked tokens require distant contextual information for reconstruction. To reduce destructive drift from the original pretrained parameters during context extension, we apply Elastic Weight Consolidation (EWC)~\cite{kirkpatrick2017overcoming} regularization with $\lambda = 1000$. Training uses a conservative fine-tuning regime with learning rate $1\times10^{-5}$ and a \texttt{constant\_with\_warmup} schedule with warmup ratio 0.1. We train the extended encoder for a single epoch.

\textbf{Hallucination detection} is formulated as token classification. The input sequence is constructed as $\text{Input}=[\text{Context}]\oplus[\text{SEP}]\oplus[\text{Query}]\oplus[\text{SEP}]\oplus[\text{Response}]$. Supervision is applied only to tokens in the response span, while tokens belonging to the context and query are masked out from the loss (assigned label \texttt{-100}). The model is fine-tuned on a mixture of the RAGTruth, DART, and E2E datasets, for a total of 21,290 training samples. We use standard \texttt{CrossEntropyLoss} without class weighting. In preliminary experiments we found that focal loss or heavy class reweighting increases recall but substantially harms precision by over-predicting hallucinations. Training uses a learning rate of $1\times10^{-5}$ with a \texttt{constant} schedule for 6 epochs and batch size 32.

\textbf{Early-exit adapters} are trained on top of the frozen hallucination detection backbone. Each adapter uses a bottleneck dimension of 256 and contains approximately 265K parameters. Training uses a combined supervised and distillation objective to approximate the behavior of the full-depth model. The loss function is
\begin{equation}
\mathcal{L} = 0.5\,\mathcal{L}_{\mathrm{CE}} + 0.5\,\mathcal{L}_{\mathrm{KL}},
\end{equation}
where $\mathcal{L}_{\mathrm{CE}}$ is the token classification loss and $\mathcal{L}_{\mathrm{KL}}$ performs distillation from the teacher model with temperature $T=2.0$. Adapter parameters are optimized using a learning rate of $2\times10^{-4}$ for 6 epochs.

\section{Deployment Recommendations}
\label{app:deployment}
Section~\ref{sec:early_exit} discusses the trade-offs between early-exit layers and full-depth inference. In practice, deployment decisions depend on the latency requirements of the application as well as the typical sequence length of the input. Because attention complexity grows quadratically with sequence length, the relative benefit of early exit increases for long-context inputs. For many real-world RAG verification pipelines, intermediate exit layers provide a favorable balance between accuracy and efficiency, allowing substantial reductions in compute while maintaining most of the performance of the full model.

In production settings, we generally recommend using an intermediate exit configuration for both real-time verification and long-document processing. This configuration preserves most of the accuracy of the full model while substantially reducing inference cost. The full-depth configuration is most appropriate when the highest possible accuracy is required, for example in offline evaluation or quality-critical analysis pipelines.

Table~\ref{tab:recommended_configs} summarizes several recommended configurations corresponding to common deployment scenarios. These configurations instantiate the fast, balanced, and high-recall operating points described in Section~3.2.5 and illustrate the resulting throughput and latency trade-offs.

\begin{table}[t]
\centering
\caption{Recommended configurations by use case.}
\label{tab:recommended_configs}
\begin{tabular}{l l r r}
\toprule
Use Case & Configuration & Throughput & Latency \\
\midrule
Real-time API & L16, BS=8, $\leq$2K & 685/s & 1.5ms \\
Batch processing & L22, BS=8, $\leq$2K & 487/s & 2.1ms \\
Long documents & L16, BS=4, 16K & 43/s & 24ms \\
Max accuracy & L22, BS=1, 24K & 15/s & 63ms \\
\bottomrule
\end{tabular}
\end{table}

A practical rule of thumb is to use the L16 configuration as the default operating point. This exit layer retains most of the accuracy of the full model while reducing inference cost, especially for longer sequences where attention dominates compute. When throughput is the primary constraint and inputs are short, pairing L16 with larger batch sizes enables real-time verification with sub-millisecond latency. For long documents, smaller batch sizes help maintain memory headroom while keeping latency manageable. The full-depth configuration (L22) is typically reserved for scenarios where maximum accuracy is required and additional latency is acceptable.

\section{Complete Latency Results}
\label{app:latency}
To provide a more complete view of the efficiency characteristics of the proposed approach, we report full throughput measurements across sequence lengths and batch sizes. All measurements are obtained using FlashAttention-2~\cite{daoflashattention} on AMD Instinct MI300X.

The results highlight several consistent trends. For short sequences up to approximately 2K tokens, batching significantly improves throughput, reaching more than four times the single-sample throughput in some settings. As sequence length increases, memory bandwidth and quadratic attention costs begin to dominate, and the throughput benefit of larger batch sizes diminishes. At very long contexts (16K–24K tokens), throughput becomes largely memory-bound and remains relatively stable across batch sizes.

These results also illustrate why early-exit strategies become increasingly beneficial at long sequence lengths. Because each transformer layer incurs quadratic attention cost, skipping later layers can substantially reduce total compute, producing larger relative speedups than in short-context scenarios.

\begin{table}[t]
\centering
\caption{Complete throughput at full depth (L22) in samples/sec using FlashAttention-2 on AMD MI300X.}
\label{tab:latency_full}
\begin{tabular}{lrrrr}
\toprule
SeqLen & BS=1 & BS=2 & BS=4 & BS=8 \\
\midrule
512  & 62 & 130 & 254 & 487 \\
1K   & 63 & 125 & 245 & 499 \\
2K   & 62 & 125 & 250 & 471 \\
4K   & 62 & 120 & 207 & 222 \\
8K   & 59 & 81  & 86  & 92  \\
16K  & 28 & 30  & 32  & 31  \\
24K  & 15 & 16  & 16  & 16  \\
\bottomrule
\end{tabular}
\end{table}

\end{document}